\documentclass[conference]{IEEEtran}
\IEEEoverridecommandlockouts
\usepackage{cite}
\usepackage{amsmath,amssymb,amsfonts}
\usepackage{algorithmic}
\usepackage{graphicx}
\usepackage{textcomp}
\usepackage{xcolor}
\def\BibTeX{{\rm B\kern-.05em{\sc i\kern-.025em b}\kern-.08em
		T\kern-.1667em\lower.7ex\hbox{E}\kern-.125emX}}
\begin{document}
	
	\title{Evaluating Empathy in Artificial Agents\\
	}
	
	\author{\IEEEauthorblockN{{\"O}zge Nilay Yal{\c c}{\i}n}
		\IEEEauthorblockA{\textit{School of Interactive Arts and Technology}\\
			\textit{Simon Fraser University}\\
			Vancouver, BC, CANADA \\
			oyalcin@sfu.ca}
	}
	
	\maketitle
	
	\begin{abstract}
		The novel research area of computational empathy is in its infancy and moving towards developing methods and standards. One major problem is the lack of agreement on the evaluation of empathy in artificial interactive systems. 
		Even though the existence of well-established methods from psychology, psychiatry and neuroscience, the translation between these methods and computational empathy is not straightforward. It requires a collective effort to develop metrics that are more suitable for interactive artificial agents. This paper is aimed as an attempt to initiate the dialogue on this important problem. We examine the evaluation methods for empathy in humans and provide suggestions for the development of better metrics to evaluate empathy in artificial agents. We acknowledge the difficulty of arriving at a single solution in a vast variety of interactive systems and propose a set of systematic approaches that can be used with a variety of applications and systems. 
	\end{abstract}
	
	
	\section{Introduction}
Emerging technologies continue to change the ways in which we interact with computers. Computational systems are evolving from being mere tools to assistants, trainers and companion agents. All of these new roles assigned to these systems highlight the importance of embodying these agents with social and emotional capabilities. The advances in computational interaction techniques allowed for the development of emotionally sensitive, perceptive, socially situated and expressive agents. One of the novel and exciting addition to these behaviors is empathy, as a complex socio-emotional behavior.

Empathy can be defined as the capacity to perceive, understand and respond to others' emotions in a manner that is more suitable to those perceived emotions than one's own \cite{preston2002empathy}. The long history of empathy research with the contribution of many disciplines (e.g. philosophy, psychology, neuroscience, ethology) resulted in a diverse set of definitions and behaviors assigned to empathy (see \cite{coplan2011empathy} for the history of the field). Behaviors such as mimicry, affective matching, consolidation and perspective-taking are assigned to empathic ability in humans \cite{coplan2011understanging}, which are crucial to initiating and maintaining social relationships. Following the centuries-old research that shows the importance of empathy in social interactions, computational empathy emerged as a novel field to equip artificial agents to show empathic behavior during their interactions.

With the recent developments in the capabilities of computational systems, the computational modeling of empathy has gained increasing attention in the last decade. Research on computational modeling of empathy have shown that empathic capacity in interactive agents lead to more trust \cite{brave2005computers, leite2014empathic}, increase the length of interaction \cite{bickmore2005establishing}, help coping with stress and frustration \cite{prendinger2005using} and increase engagement \cite{leite2014empathic} (see \cite{paiva2017empathy} for a review of the field). These findings suggest that agents with empathy could enhance the social interaction in educational applications, artificial companions, medical assistants and gaming applications. Equipping artificial social agents with empathic capabilities is, therefore, a crucial and yet challenging problem. 

Part of this challenge arises from the lack of evaluation methods to measure empathic behavior in artificial agents, which would allow for a systematic assessment of the steps need to be taken to model the components of this complex socio-emotional phenomenon. 
Although various fields provide well-established evaluation methods to measure empathy in humans, it is not clear how to translate these methods to evaluate computational systems. Empathy research in psychology provides validated methods to measure empathy levels in a person \cite{davis1983measuring, baron2004empathy, lawrence2004measuring}. These methods often require a first-person report of certain behavioral traits based on subjective questionnaires. This poses a challenge for artificial empathy work in virtual agents, as subjective measurements cannot be used in artificial entities \cite{paiva2017empathy}. Moreover, behavioral tests from psychology and neuroscience often rely on physiological signals such as neural activity, heart rate and skin conductance, which cannot be used for machines. Other methods may require the observation of experts in the field, which is hard to automate.

Furthermore, agents that have varying levels of interactive capabilities and application goals can have different effects on the perception of the agent and interaction. The characteristics of the agent (e.g. aesthetics, embodiment) as well as non-functional properties (e.g. fluency, response time) can affect the evaluation of empathy, as much as the empathic functionality. A conversational agent in a text-only environment such as chatbots would require emotion perception and expression using different modalities than an embodied conversational agent. In that sense, every additional capability of the agent would contribute to the evaluation of the interaction and the system. Similarly, the application areas may enforce varying levels of expectations in terms of empathic behavior. A medical assistant may require more sympathy, where a personal trainer would focus on pushing boundaries of the interaction partner. This diversity highlights the importance of a set of evaluation methods that allow for flexibility, instead of focusing on reaching to a single solution.

This paper aims to provide recommendations on how to systematically evaluate empathy in artificial agents in a variety of contexts and capabilities. We propose to approach this problem by focusing on best practices in both the empathy research in humans and the human-computer interaction (HCI) research. To achieve this, we will examine the methods of evaluation in empathy of humans to draw conclusions on how to adapt the key concepts to computational empathy. We propose to use system-level and feature-level evaluations to systematically list the factors that contribute to the evaluation of empathy. By providing a checklist of these factors, we aim to initiate the discussion towards creating a common ground on the evaluation of empathy in artificial agents. 

In the following sections, we will explore the evaluation methods on empathy in humans (Section \ref{sec:empathy}) and try to apply this know-how into evaluating empathy in artificial agents using system-level and feature-level evaluations (Section \ref{sec:agents}). We will conclude with a discussion of challenges and a call for collective action to work towards new evaluation methods in this new and exciting field.

\section{Evaluation of Empathy in Humans}
\label{sec:empathy}
Empathy research from many disciplines have developed definitions and models for empathy that resulted in a variety of capabilities assigned to empathic behavior \cite{coplan2011understanging}. Capabilities such as mimicry, affective matching (emotional contagion), sympathy (empathic concern), altruistic helping (consolidation) and perspective taking are assigned to empathic behavior by scholars \cite{coplan2011empathy, de2017mammalian, batson2009}. How many of these behaviors should constitute empathic behavior and how they are connected are still highly debated topics in empathy research. Following the differences in the definitions and models of empathy, the evaluation metrics developed to measure empathic behavior may vary dramatically. %According to the chosen definition and model, different evaluation metrics proposed.

Some definitions of empathy separate a number of these capabilities as affective and cognitive empathy \cite{omdahl2014cognitive}. According to this view, affective empathy refers to the relatively automatic emotional responses to other's emotions. Behaviors related to affective empathy can be listed as mimicry, affective matching and empathic concern \cite{1987Eaid}. On the other hand, cognitive empathy includes behaviors that require the understanding of another's emotional state and behaviors. Behaviors such as consolidation, perspective-taking and altruistic helping are said to originate from the involvement of cognitive mechanisms during the processing of the other's emotional situation \cite{de2017mammalian}.

Contrary this dual view of empathy that separates affective and cognitive processes, a unifying view of empathy is gaining attention as an alternative. These recent models and definitions of empathy suggest a more multi-dimensional approach where both affective and cognitive empathy are interconnected with a variety of processes that results in individual differences \cite{davis1980multidimensional, preston2002empathy, hoffman2001empathy}. One of the most prominent views of empathy called the Russian Doll Model of Empathy \cite{de2007russian} suggests hierarchical levels of affective and cognitive capabilities are connected through evolutionary mechanisms. According to this model, processes such as emotional communication capabilities (recognition and expression), emotion regulation, appraisal processes and theory of mind are considered to be the foundational mechanisms that allow the levels of affective and cognitive empathic behavior \cite{de2017mammalian}. It is also suggested that the individual differences between the empathic responses of people are related to the factors that affect the outcome of these processes \cite{davis1980multidimensional}. For instance, the recognition of emotions depends on the intensity of the perceived emotion, where the regulation of the emotion would depend on the familiarity between individuals as well as the features of the observer (mood, personality) \cite{de2006empathic}. 

The evaluation metrics to measure empathic capacity in humans varies depending on the definition and the capabilities that are assigned to empathic behavior. Evaluation metrics that follow the categorical view of empathy usually assess the affective or the cognitive aspects of empathic behavior as separate constructs. On the other hand, evaluation metrics that follow the multi-dimensional view focuses on defining and evaluating the levels of processes and behaviors that determine the extent of empathic behavior. Although both of these approaches received criticism by some researchers that suggest empathy should only involve the higher level processes \cite{coplan2011understanging}, it is useful to focus on a broader view of empathy for to arrive at a comprehensive framework to evaluate empathy in artificial agents. Following this notion, we will explore the evaluation methods by focusing their adaptability to computational empathy research.

In this section, we will give an overview of the well-established measurements for empathic behavior, while categorizing them in terms of the focus of empathic behavior and the method of delivery.  
Evaluation metrics can target a variety of levels of empathic behavior with different levels of granularity and abstraction. 

Some of the evaluation methods are designed to measure empathy as a single comprehensive construct and aimed to derive a single value that would indicate the global empathy. Others focus on multiple features or a subset of those features that underlie empathic behavior, such as the affective and cognitive capabilities that are mentioned earlier. Therefore, we will categorize the evaluation metrics according to the level of granularity they are aimed to quantify: global empathy and components of empathy.

Moreover, the method of delivery for these evaluations can be categorized as self-report, observational and physiological approaches. Physiological approaches include measurements of brain activity or autonomic nervous system measures (heart rate, skin conductance, breathing rate). Methodologically, there is no direct way of applying the physiological approaches to virtual agents. Therefore, we will not cover them in our paper (see \cite{NeumannDavid.L.2015C1-M} for a review of these approaches). Self-report measures usually include surveys/questionnaires that rely on the individual's assessment of their behavior. Observational methods can include the behavioral tests and perceived empathy measures.
Behavioral methods rely on performance tests based on experimental stimuli. These methods are often used to assess the components of empathy in humans and aimed to indicate deficiencies.  
Lastly, perceived empathy metrics are questionnaires that require an observer's assessment of an individual's behaviors. These can include expert observations on the subject's behaviors as well as a second or third person account of a non-expert.
In the following sub-sections, we will give detailed examples on the well-established evaluation metrics on each of these methods within global and component-based evaluation metrics.

\subsection{Evaluating Global Empathy}
Measures of global empathy are aimed at quantifying a single value that would indicate the strength of empathic capability as a broader concept.
Many researchers have attempted to develop self-report measures of empathy based on the definitions and capabilities they assigned to the term. Most of these methods focus on the evaluation of empathy as a whole, while others focus on the specific factors that add up to the global empathic behavior. 

One of the earliest self-measure of empathy is Hogan's Empathy Scale (ES) \cite{hogan1969development} that is mostly used to assess cognitive empathy with 64 true-false statements taken from the standard psychological scales. This questionnaire was intended to examine the relation of empathy with moral and socially appropriate behavior. It was criticized by later works that it is better suited for the evaluation of social skills in a broader sense rather than a specification of empathy \cite{davis1983measuring, baron2004empathy}. Moreover, the low scores of the validity and reliability of the scale resulted in a continuous decrease in the use of this scale as a valid measure of empathy \cite{froman2001rethinking}. However, this attempt encouraged researchers to investigate further and examine the development of a more suitable evaluation of empathy.

A frequently used example is Davis's Interpersonal Reactivity Index (IRI) \cite{davis1983measuring}, is a 28-item scale for multi-dimensional measurement of empathy with four sub-scales: perspective-taking, empathic-concern, fantasy and personal distress. However, there have been some discussions around the appropriateness of this scale to measure empathy. Firstly, it was argued that the questionnaire may capture behaviors broader than empathy \cite{baron2004empathy, spreng2009toronto}, such as imagination (e.g. item 1 "I daydream and fantasize, with some regularity, about things that might happen to me") and emotional control (e.g. item 10 "I sometimes feel helpless when I am in the middle of a very emotional situation"). An adaptation of IRI to exclude the ``fantasy" subscale was later adopted as Feeling and Thinking Scale \cite{garton2005development}. It was also suggested that the ``personal distress" sub-scale mostly measures anxiety towards distressing situations in general and does not relate to the core functions of empathy. Moreover, some researchers suggested the further refinement of this scale due to the correlation between these sub-scales \cite{spreng2009toronto}.

The Empathy Quotient (EQ) \cite{baron2004empathy} is one of the most accepted self-report scales that is validated by numerous studies \cite{lawrence2004measuring}. Authors define empathy as ``the drive to identify another person's emotions and thoughts and to respond to these with approapriate emotion" (p.361). This test is aimed as a clinical screening tool for adults with Autism Spectrum Disorders. In contrast with other self-report questionnaires of empathy, authors did not differentiate between affective and cognitive empathy. They aim to capture empathy in a broader sense where both levels have very interrelated capacities. This questionnaire includes 60 items with 40 empathy-related and 20 filler questions answered with a 4-point Likert scale that scores agreement with the statements. Example questions from the EQ are ``I am good at predicting how someone will feel" and ``Seeing people cry doesn't really upset me." The questionnaire scores are shown to correlate with autism and gender differences \cite{baron2004empathy}.

A recent attempt to further examine and combine these self-report measures uses factor-analysis to reach to a brief and reliable measurement of empathy is called The Toronto Empathy Questionnaire \cite{spreng2009toronto} (TEQ). Authors gathered a total of 142 items from several self-report empathy questionnaires such as IRI, ES, BEES, QMEE, AQ as well as empathy questionnaires for specific populations such as Jefferson Scale of Physician Empathy \cite{hojat2001empathy}, Nursing Empathy Scale \cite{reynolds2017measurement} and Japanese Adolescent Empathy Scale \cite{hashimoto2002structure}. Authors used these items to refine a final set of 16 items that are found to be most correlated with Empathy scores compared to other questionnaires. Responses are made with 5-point Likert scale items that show agreeableness of the statements. This questionnaire is a shorter alternative to the EQ with high internal consistency, validity and reliability scores. 

A similar approach is taken in the Questionnaire of Cognitive and Affective Empathy (QCAE) \cite{reniers2011qcae}, which is derived from EQ, ES, IRI and the Impulsiveness-Venturesomeness-Empathy Inventory \cite{eysenck1978impulsiveness}. Authors finalized a 31-item questionnaire that measures cognitive and affective empathy, as the name suggests. The QCAE consists of five sub-scales, where two sub-scales are related to cognitive empathy (perspective-taking, online sumulation), and three of them are related to affective empathy (emotion contagion, proximal responsivity and peripheral responsivity). 

Other methods focus on the evaluation of the perception of empathic behavior. These measurements provide a second and third person perspective on an individual's empathy with questionnaires. Jefferson Scale of Physician Empathy is developed to evaluate empathy as a predominantly cognitive attribute \cite{hojat2001empathy}. These components are ``communication", ``understanding" and ``cognition", are focused on the cognitive empathy, rather than the affective empathy that was mentioned earlier (see Section \ref{sec:empathy}). The scale consists of 20 items with a 7-point Likert type scale on agreement with the statements. %The items
Another example for the perceived empathy methods is the Consultation and Relational Empathy (CARE) Questionnaire \cite{mercer2004consultation}. CARE was developed to measure ``relational empathy" that focuses on the social function of empathy. It consists of 10 statements that start with ``How was the doctor at ..." and scored by the patient by using a 5-point likert scale from ``poor" to ``excellent". The items include ``Fully understanding your concerns", ``Showing care and compassion" and ``Being positive". Some of these items show significant overlap with self-report questionnaires such as the IRI.

\subsection{Evaluating Components of Empathy} 
An alternative approach to the evaluation of global empathy is the evaluation of specific components that are required for empathic behavior. These evaluations are usually done by testing behavioral and cognitive abilities to detect the deficits and abnormalities in the behavior. Components such as emotional communication (recognition and expression), emotion regulation, appraisal and perspective taking are usually targeted in these evaluations as critical mechanisms for levels of empathy. 

The ``reading the mind in the eyes" test \cite{baron1997another} was one of the first examples of these behavioral tests. This test was aimed to be used as a screening test for adults or children with Aspergers Syndrome, who are considered to have a deficit in empathy. The revised version of this test consists of 36 photographs of the eye-region of the face that shows different emotional expressions \cite{baron2001reading}. The participants are presented with these photographs and are asked the most appropriate word to describe ``what the person in the photograph is thinking or feeling" (p.241) among four words that are presented. The target terms include words that show mental states such as ``thoughtful", ``interested" or ``fantasizing", as well as words that relate to the emotional state such as ``upset", ``nervous" or ``hostile". The test results indicate the ability to perceive social and emotional cues where a lower score is associated with a broader set of phenomena than just measuring empathy. 

Similarly ``reading the mind in the voice" test \cite{golan2007reading} and ``reading the mind in films" test \cite{golan2006reading} are aimed to measure the ability to detect socio-emoitonal cues in voice and movie stimuli respectively. The ``voice" test uses segments of dialogue taken from dramatic performances, where the ``films" test uses audio-visual recordings from movies that shows complex situations. These behavioral tests target the Theory of Mind (ToM), that is the ability to attribute mental states (beliefs, desires, intentions and emotions) to others that are distinct to one's own. ToM being a crucial part of cognitive empathy, these simple tests allow to spot deficiencies while targeting necessary perceptual abilities.

Understanding appraisals \cite{baron1986mechanical}, \cite{lawson2004empathising} and intuitive physics \cite{baron2001intuitive} can also be used to determine the capacity to understand cause and effect relationships, which is based on the higher level cognitive mechanisms. The Picture-Stories task \cite{baron1986mechanical} consists of a series of pictures that show the cause and effect relationship in social situations when appropriately sequenced. Similarly, the Social Stories Questionnaire (SSQ) \cite{lawson2004empathising} consists of 10 short stories that may involve situations where one character could upset the other character in the story. Participants are asked to whether a selected utterance from the story contains an upsetting utterance and whether the behavior of one character could have upset the other character. The number of correct answers defines the SSQ score in this test.

\section{Evaluation of Empathy in Interactive Agents}
\label{sec:agents}
Being a novel field, empathy studies in artificial intelligence (AI) has no strong standardization and validated methods to measure empathy in artificial agents. In the previous section, we laid out some of the most accepted evaluation methods to evaluate empathy in humans. Although these methods are well-established and agreed upon in the academic community, applying them in the context of artificial agents is not straightforward. Most of these tools rely on self-measurement which cannot be applied to computational systems or the assessment of an expert that is difficult to automatize. Moreover, the differences in the capabilities of the agents and the application context restrict the usage of general behavioral measurements. These issues made it challenging to use this know-how to the evaluation of empathy in artificial interactive agents.

Empathy measurements in psychology literature include the evaluation of specific cognitive and behavioral capabilities as well as an overall evaluation of empathy.  
Specific features include evaluations of emotion recognition \cite{baron2001reading}, perspective taking \cite{davis1980multidimensional} and empathic concern \cite{davis1983measuring}. Understanding the user's emotion depends on the correct recognition of the facial expressions and the performance of the emotion classifier. The perception of empathic behavior depends on the successful expression of the intended empathic emotion. Overall evaluation of empathy should take these feature's performance along with the system-level evaluation of empathy. 

Similarly, the performance of computational systems highly depends on the performance and accuracy of the individual components as well as the integration of these components at the system-level. Due to the complexity and multi-component nature of interactive agents, scholars suggested \cite{dybkjaer2004evaluation, ruttkay2006evaluating} to provide feature-level and system-level evaluations separately. System-level evaluations focus on the behavior of the agent as a whole, where feature-level evaluations are aimed to isolate individual components of the system separately.

Following this notion, we propose to combine the best practices in the HCI research with the traditional methods of evaluating empathy. In the following sub-sections, we will focus on how to evaluate empathy using system-level and feature-level evaluation methods. We will systematically list the factors that contribute to the evaluation of empathy in artificial agents to initiate the discussion towards creating a common ground. 

\subsection{System-Level Evaluation}
System-level evaluation is focused on the measurement of the behavior of the system in a broader sense. Similar to the self-report and perceived empathy evaluations that are aimed at capturing the global empathic behavior, system-level evaluations in artificial agents focus on the overall perception of empathy of the agent. In these type of evaluations, the participants interact with the complete system according to the interaction context, and a set of subjective and objective evaluations are used to compare the different versions of the system or with human behavior.

Previous studies in artificial empathy often focus on the second person or third person perception of empathy of the systems by using empathy-related terminology such as ``feeling with" \cite{rodrigues2015process}, ``feeling with" \cite{boukricha2013computational}, ``emotion matching" \cite{mcquiggan2008modeling}, ``compassionate" \cite{ochs2012formal} or ``caring" \cite{brave2005computers}.  However, these terms only focus on one specific aspect of empathic behavior or related constructs. 
An interesting approach was used to train and evaluate the CARE framework \cite{mcquiggan2008modeling} by comparing the behavior of the agent with human behavior in a goal-directed environment. This approach can be automated but requires additional data-collection and evaluation steps of human behavior in a similar context to allow for a direct comparison. 

As computational empathy research gaining more attention, researchers are beginning to raise awareness on the importance of using more suitable metrics. Recently scholars \cite{paiva2017empathy} suggested using a variation of the IRI questionnaire \cite{davis1983measuring} by adapting the first-person evaluation to a perceived-empathy survey. This idea was applied as a part of the EMOTE project \cite{barendregt2016} authors assessed the perceived empathy of a social robot using the IRI questionnaire. Similarly, Toronto empathy questionnaire \cite{spreng2009toronto} was used as a perceived empathy metric by converting the self-report questionnaire into a second or third person evaluation \cite{yalcin2019evaluating}. 
These evaluation methods can be used to evaluate the system by the interaction partner using a questionnaire. Moreover, the evaluation can be done by a third-person after watching the live or recorded interaction between the system and a participant.
Although these methods provide an evaluation that is aligned with the related research on empathy, they were not validated.

However, these perceptual evaluations of empathy can be affected by several factors that should be taken into consideration while applying these system-level evaluations. These factors can be categorized as user-related factors, context-related factors and system-related factors.

\subsubsection{User-related Factors} Research on empathy shown that humans empathize with each other on different levels depending on factors such as their gender, mood, personality, similarity and social capabilities \cite{de2006empathic, davis1980multidimensional, hoffman2001empathy}. These findings highlight the importance of controlling for these factors in a comparative evaluation of the agent behavior. Moreover, individual traits such as culture, socio-economic background and computer experience might affect the evaluation of the system as an interactive tool \cite{reeves1996media}. 

\subsubsection{Context-related Factors} Relationship and context related factors would impact the strength and expression of empathic behavior. The context, the appraisal of the situation or the social role of the empathizer are suggested to influence the regulation of emotions \cite{omdahl2014cognitive}. Systems that act as companions as opposed to trainers are expected to be more friendly. This user expectancy based on the role of the agent and the context can effect the perception of empathy, where people tend to be more empathic towards in-group members such as friends and family members \cite{hoffman2001empathy, de2006empathic}. Moreover, goal-directed factors that show the quality of experience such as effectiveness, efficiency, user-satisfaction, utility and acceptability can influence the overall perception of the system \cite{ruttkay2006evaluating}. 

\subsubsection{System-related Factors} Factors related to the system behavior that are not directly linked to its empathic capacity can also impact the evaluation of empathy. Studies have shown that aesthetic characteristics of the interaction partner have a dramatic influence on the perception of empathy in humans \cite{muller2013empathy}. These aesthetic considerations might translate into the factors related to the looks, human-likeness, fluency of movement and believability of the agents \cite{misselhorn2009empathy, loyall1997believable}. HCI research has developed evaluation metrics to control the effect of these factors such as anthropomorphism, animacy, likability, perceived intelligence and perceived safety \cite{bartneck2009measurement}. 

Computational empathy research has already been measuring some of these factors as control variables as well as additional metrics for the overall success of their system \cite{ochs2012formal, rodrigues2015process, leite2014empathic}.
Although the effects of the factors related to empathic behavior are examined in detail in empathy research, the relationship between these factors and the perception of empathy is yet to be examined.

\subsection{Feature-Level Evaluation}
In addition to the system-level evaluation, the evaluation of individual aspects of the system is necessary to assess the empathic capabilities of an interactive system. Feature-level evaluations can provide an incremental assessment of each component and capability of the agent. This allows for capturing the propagation of errors in empathic behavior, similar to the behavioral evaluations in empathy research that focuses on capturing deficits in empathic capacity.

In complex interactive systems, the evaluation methodologies usually include the metrics from various sub-fields, such as speech recognition, emotion recognition and speech synthesis \cite{ruttkay2006evaluating}. The performance of the implementation is depended on the success of the separate features of the system, as each component affects the evaluation of other components. Therefore, the deficits in one capability might drastically influence the other. For example, the appraisal mechanism could be effected by simply a poorly performing emotion recognition component. Similarly, according to the empathic capabilities implemented to an agent, the features of every capability should be evaluated systematically at every stage of development.

For the evaluation of empathy as a broader concept, we will use the categorization of empathy features based on the evolutionary approaches \cite{de2007russian, de2017mammalian} as we discussed in Section \ref{sec:empathy}. According to these approaches, the empathic capacity can be categorized into three hierarchical mechanisms: emotional communication, emotion regulation and cognitive processes. Similar components have been proposed by other researchers in empathy \cite{paiva2017empathy, yalcin2018computational} and emotional intelligence research \cite{Scherer2007emotional}. However, it should be noted that different types of definitions, models of empathy, as well as the capabilities and goals of interactive agents would require the evaluation of different subsets of these capabilities.

\subsubsection{Emotional Communication}
Emotional communication capacity forms the foundation of affective behaviors including empathy \cite{Scherer2007emotional, de2017mammalian}. This capacity can be further categorized as emotion recognition and emotion expression components. The successful detection and recognition of the input emotions would directly impact the empathic behavior of every level, hence the perception of empathy of the agent. Similarly, as the empathic behavior is essentially an emotional response to the stimuli, the emotional expression ability of the agent would directly influence the empathic behavior and the evaluation of the behavior. Therefore, it is crucial to include the individual evaluations of emotional communication capacity to assess the empathic capabilities of an interactive agent.

The evaluation of the emotion recognition ability may include a variety of well-established tests depending on the input modalities of the agent. For example, a text-based conversational agent's emotional communication capability can only be tested via the text-based linguistic emotional recognition and expression, where an embodied conversational agent should also be evaluated according to its speech, body gestures and facial expressions. Similarly, the success of the emotion expression behavior should be evaluated depending on the output modalities of the agent that are going to be used for expressing the empathic emotions. Following the behavioral metrics for empathy that are designed to evaluate the emotion recognition from pictures \cite{baron2001reading}, voice \cite{golan2007reading} and complex emotions from movies \cite{golan2006reading}, the metrics for the recognition of agents should include the evaluation of each modality. Affective computing research provides well-established evaluation metrics for emotion recognition in computational systems \cite{scherer2010blueprint}.

\subsubsection{Emotion Regulation} A variety of models on empathy and emotional intelligence assign central importance in the ability to regulate emotions based on a variety of dynamics \cite{Scherer2007emotional, de2017mammalian}. Emotion regulation capacity can be based on personality and mood of the individual that allows for automatic regulation \cite{scherer2010blueprint}. Humans are found to automatically assign attributes such as personality, gender and mood to interactive systems \cite{reeves1996media}. Personality metrics such as the Big Five are widely used in affective computing research \cite{vinciarelli2014survey}. Subjective evaluation metrics for emotional control have been proposed \cite{gullone2012emotion,preece2018measuring}. However, these approaches have not been used by the empathic computing research and may require adjustments.

\subsubsection{Cognitive Processes} The higher level of empathic capacity is suggested to include the cognitive processes such as appraisal, re-appraisal, self-oriented perspective taking and other-oriented perspective taking behaviors \cite{de2017mammalian}. These cognitive processes would also control the emotion regulation abilities that allow for suppression or enhancement of emotions based on the context \cite{scherer2010blueprint}. As we covered in Section \ref{sec:empathy}, behavioral measures such as understanding appraisals \cite{baron1986mechanical}, the picture-stories task \cite{baron1986mechanical} and the Social Stories Questionnaire (SSQ) \cite{lawson2004empathising} are used to assess the deficiencies in the cognitive empathy. However, there are no standardized method to evaluate these capabilities in artificial agents. Moreover, the domain dependence and the problem of scalability for the cognitive capabilities makes it problematic to perform these tests to interactive agents with various capabilities.

Even though we suggested solutions for the feature-level evaluations to adopt the existing metrics, most of them needs further adjustments and validations to be applied in artificial agents. 

\section{Concluding Remarks}
Empathy as a complex socio-emotional phenomena where the variety of definitions and models in the research community makes it problematic to evaluate and compare the implementation of the behavior in interactive artificial agents. This article is aimed to describe in detail the methods have been developed in the empathy research to evaluate empathic behavior, that can be translated into the emerging computational empathy research. We attempted to provide a systematic approach to the evaluation of this complex by suggesting the approach the evaluation on system-level and feature-level. As we acknowledge the difficulties of establishing a common ground in a diverse set application areas and capabilities of agents, we believe the importance of specifying the broader picture in the evaluation of empathy. 

Our goal was to provide a guide on how we can evaluate the empathic behavior in artificial agents. We proposed system-level and feature-level evaluations for computational empathy systems to approach the issue systematically. We further provided a list of factors and components that can be used as a road-map to create individual evaluations for empathic systems in various application areas. We propose that the extensive body of work in the evaluation of empathy in humans, and the evaluation methods from affective and social computing can be used for computational empathy research.  
We hope to initiate the discussion towards creating a common ground to evaluate and compare computational empathy methods with this paper. We believe that a collective effort is required to develop specific measures and evaluation frameworks of empathy for interactive artificial agents.

	\bibliographystyle{IEEEtran}
	\bibliography{IEEEabrv,sample-bibliography}

\end{document}